\documentclass[runningheads]{llncs}
\usepackage{url}
\usepackage[utf8]{inputenc}
\usepackage[small]{caption}
\usepackage{graphicx}
\usepackage{amsmath}
\newtheorem{hypothesis}{Hypothesis}

\usepackage{booktabs}
\usepackage{algorithm}
\usepackage{algorithmic}
\usepackage{graphicx}
\usepackage{comment}
\usepackage{amssymb} % define this before the line numbering.
\usepackage{gensymb}
\usepackage{tikz}
\usepackage{tabularx}
\newcolumntype{P}[1]{>{\centering\arraybackslash}p{#1}}

\usepackage{color}
\usepackage[hidelinks,breaklinks]{hyperref}
\usepackage{epsfig}
\usepackage{amstext}
\usepackage{textcomp}
\usepackage{multirow}
\usepackage{hhline}
\usepackage[misc]{ifsym}

\def\hasappendix{True}
\def\true{True}
\begin{document}

\title{Few-Shot Domain Adaptation with Polymorphic Transformers}

\author{Shaohua Li\inst{1} \and Xiuchao Sui\inst{1} \and Jie Fu\inst{2} \and Huazhu Fu\inst{3} \and Xiangde Luo\inst{4} \and Yangqin Feng\inst{1} \and  Xinxing Xu\inst{1}$^\text{(\Letter)}$ \and Yong Liu\inst{1} \and Daniel Ting\inst{5} \and Rick Siow Mong Goh\inst{1}} 

%index{Li, Shaohua}
%index{Sui, Xiuchao}
%index{Fu, Jie}
%index{Fu, Huazhu}
%index{Luo, Xiangde}
%index{Feng, Yangqin}
%index{Xu, Xinxing}
%index{Liu, Yong}
%index{Ting, Daniel}
%index{Goh, Rick Siow Mong}

\institute{
Institute of High Performance Computing, A*STAR, Singapore \\ \email{xuxinx@ihpc.a-star.edu.sg} \\
\and
Mila, University of Montreal \\
\and 
Inception Institute of Artificial Intelligence, United Arab Emirates \\
\and
University of Electronic Science and Technology of China \\
\and
Singapore Eye Research Institute
}

\titlerunning{Few-Shot DA with Polymorphic Transformers}
\authorrunning{S. Li et al.}

\maketitle

\begin{abstract}
Deep neural networks (DNNs) trained on one set of medical images often experience severe performance drop on unseen test images, due to various domain discrepancy between the training images (source domain) and the test images (target domain), which raises a domain adaptation issue. In clinical settings, it is difficult to collect enough annotated target domain data in a short period. Few-shot domain adaptation, i.e., adapting a trained model with a handful of annotations, is highly practical and useful in this case. In this paper, we propose a Polymorphic Transformer (\emph{Polyformer}), which can be incorporated into any DNN backbones for few-shot domain adaptation. Specifically, after the polyformer layer is inserted into a model trained on the source domain, it extracts a set of prototype embeddings, which can be viewed as a ``basis'' of the source-domain features. On the target domain, the polyformer layer adapts by only updating  a projection layer which controls the interactions between image features and the prototype embeddings. All other model weights (except BatchNorm parameters) are frozen during adaptation. Thus, the chance of overfitting the annotations is greatly reduced, and the model can perform robustly on the target domain after being trained on a few annotated images. We demonstrate the effectiveness of Polyformer on two medical segmentation tasks (i.e., optic disc/cup segmentation, and polyp segmentation). The source code of Polyformer is released at \url{https://github.com/askerlee/segtran}.

\end{abstract}
\section{Introduction}

% \cite{PANet}

Deep neural networks (DNNs) are notoriously fragile when being used on a domain not seen before. For example, it is common to witness $10\sim 20$\% drop of accuracy on images captured with a device different from the training images. The training images and unseen test images are referred to as the \emph{source domain} and the \emph{target domain}, respectively. Domain adaptation (DA), i.e., modifying an existing model trained on the source domain, so that it performs well on the target domain, is important for deploying DNNs for medical image tasks.

Domain adaptation is trivial if a large set of annotated data exist in the target domain. However, such annotations are usually expensive to acquire, especially for segmentation tasks. Though, it is still cheap and feasible to obtain a handful of annotations. This work focuses on doing DA on a handful of annotations, or \emph{few-shot} domain adaptation. It is a special case of semi-supervised learning. 
A large body of literature focuses on reducing the domain discrepancy by minimizing a domain adversarial loss \cite{revgrad,adda,posal,da-adv,adv-adapt-cell,DAM}. Such methods can be used both in unsupervised and semi-supervised settings. As shown in our experiments, they are helpful for DA, and are complementary to our method.

In practice, a common approach to DA is retraining the model on the mixed source and target domain data \cite{joint-train}. However, it may be suboptimal in the few-shot scenario, as the joint dataset is dominated by the source domain. Another popular approach is fine-tuning the model weights on the target-domain annotated data. In the few-shot scenario, however, updating the whole pretrained model could easily overfit the limited target-domain annotations \cite{no-forget}. 
%There are a plethora of methods for domain adaption. For example, specifically designed data augmentation policies \cite{gan-histo,adapt-nuclei} have been proved to be effective. However, data augmentation requires the augmentation operators cover the domain discrepancies, which is often not the case. 
A remedy is to minimize the modification to the pretrained weights. For instance, we could freeze the feature extractor and just fine-tune the task head. Another scheme is to introduce adaptive modules into existing models, such as the DAM module \cite{DAM}, and freeze the pretrained weights.

Along the line of adaptive module-based methods, we propose a \emph{polymorphic transformer} (polyformer). It can be inserted into a pretrained model, to take the responsibility of DA. It first extracts a set of \emph{prototype embeddings} from the source domain, which is a condensed representation of the source-domain features. On the target domain, by attending with the prototype embeddings, the polyformer dynamically transforms the target-domain features. Thanks to the projection mechanism of transformers \cite{transformer}, after merely fine-tuning a projection layer, the transformed target-domain features can be made semantically compatible with the source domain. Hence, it can achieve good DA performance even in the few-shot scenario. As a proof-of-concept, we demonstrate the effectiveness of the polyformer on a vanilla U-Net model \cite{unet}, evaluated on two cross-domain segmentation tasks: optic disc/cup area segmentation in fundus images, and polyp segmentation in colonoscopy images.

\begin{figure}[t]
\centering
\includegraphics[width=0.9\textwidth]{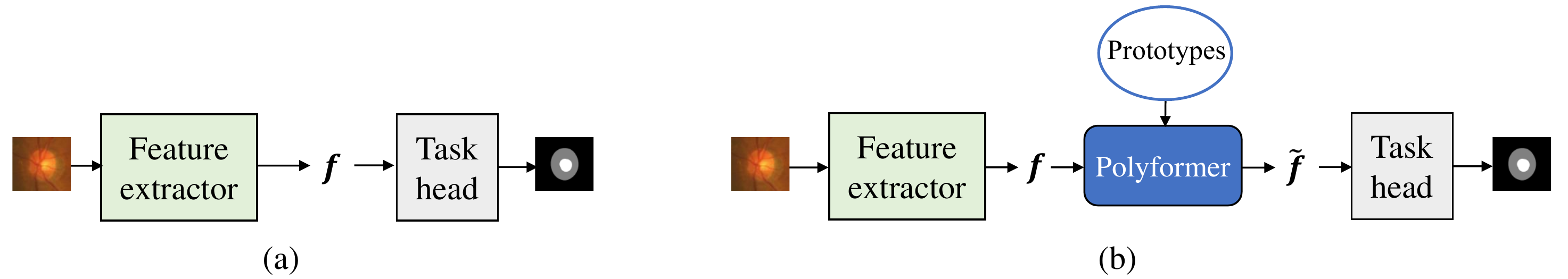}
\caption{(a) The original model pipeline consists of a feature extractor and a task head. (b) A modified pipeline for domain adaptation with an inserted polyformer. ``Prototypes'' are a set of prototype embeddings. For adaptation, the polyformer converts features $\boldsymbol{f}$ to $\tilde{\boldsymbol{f}}$. The weights of the feature extractor and the task head are frozen.} \label{pipeline}
\end{figure}

\section{Related work}
\subsubsection{Few-Shot Learning}
Few-shot learning (FSL) is closely related, but different from few-shot DA. Typically, FSL is to adapt a pretrained model, so that it performs well on \emph{novel tasks} (e.g. new classes) for which training examples are scarce \cite{oneshot-da}. In contrast, in few-shot DA, the model performs the same task on the source and target domains. A recent line of FSL research \cite{deepemd,set-to-set} is to first extract prototype embeddings (prototypes) from the support (training) images of each class, and then match them with query (test) image features using a distance metric. The prototypes are a set of ``weak classifiers'' \cite{adaboost} that vote to make the final prediction. Note that prototypes in FSL and in the polyformer serve different roles: in FSL the prototypes are used to make the final prediction, and in the polyformer they are used to transform target-domain features.

\section{The Polymorphic Transformer}

The polymorphic transformer (polyformer) is designed to bridge the gap between different domains. A pretrained network can offload DA onto a polyformer layer, so that it keeps all weights (except BatchNorms) frozen, and still performs robustly  on a new domain. Adapting a polyformer layer only requires fine-tuning a projection layer, and thus a few annotated images are sufficient. In theory, the polyformer can be incorporated with any backbone networks, such as U-Net \cite{unet}, DeepLabV3+ \cite{deeplabv3+} or transformer-based models. In this work, we choose U-Net to illustrate how a polyformer performs DA on segmentation tasks. 

Fig. \ref{pipeline} illustrates how a polyformer layer is incorporated into an existing model. In Fig. \ref{pipeline}(a), suppose a model $M$ splits into a feature extractor $M_1$ and a task head $M_2$ (A similar formulation is found in \cite{fewshot-ada}). For example, a U-Net can split into the encoder-decoder ($M_1$) and the segmentation head ($M_2$). On an input image $x$, the feature extractor generates feature maps $\boldsymbol{f}$, which are fed into $M_2$ to make predictions. On the target domain, due to domain discrepancy, the feature maps $\boldsymbol{f}$ follow different distributional properties, and thus $M_2$ is prone to make wrong predictions. 

\begin{figure}[!t]
\begin{center}
\includegraphics[width=0.6\textwidth]{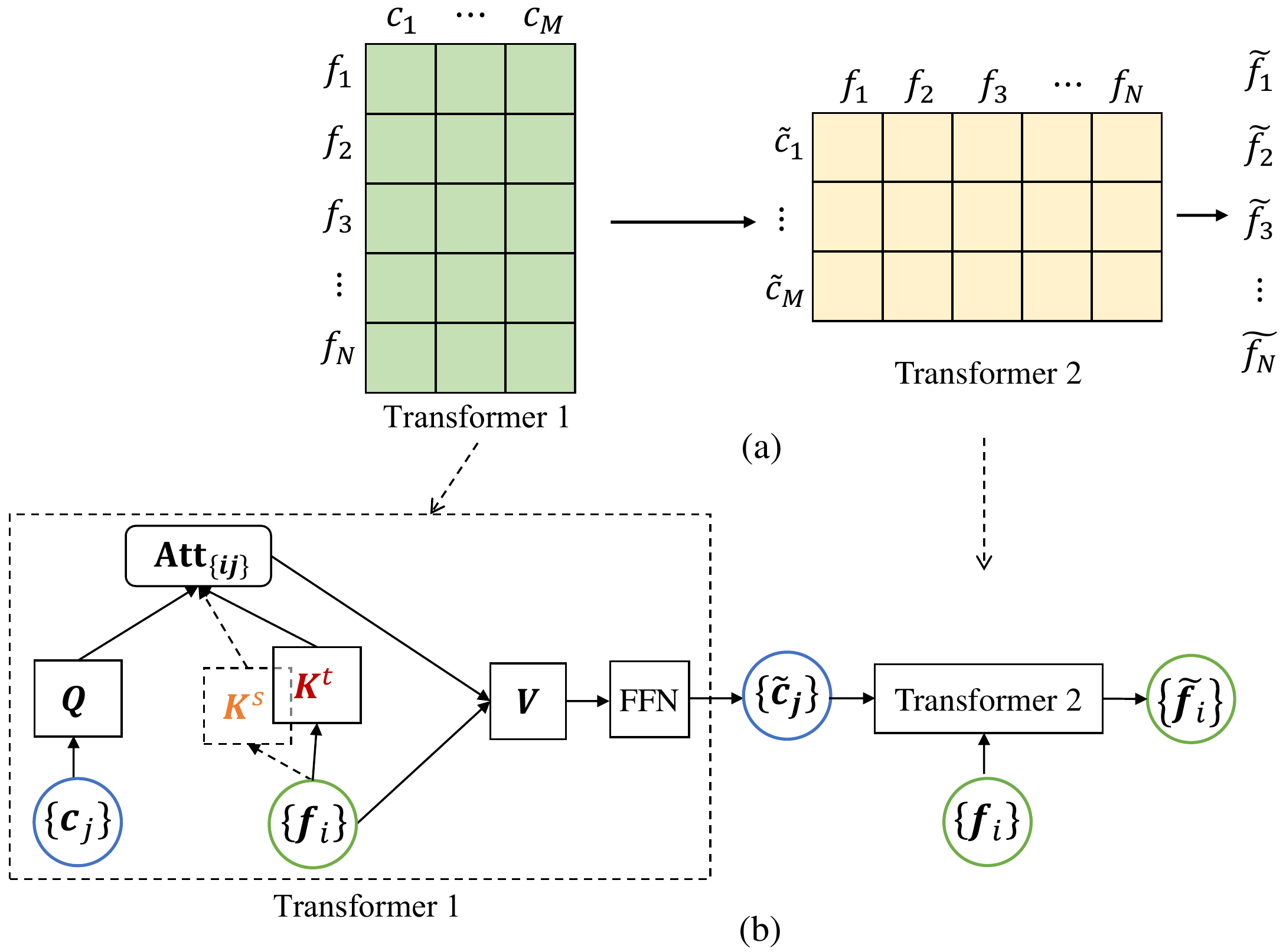}
\caption{ A polyformer layer consists of two sub-transformers 1 and 2. (a) Schematic of the polyformer attention. In Transformer 1, the input feature vectors $f_1,\cdots,f_N$ attend with the prototypes $\boldsymbol{c}_1,\cdots,\boldsymbol{c}_M$, yielding  $\tilde{\boldsymbol{c}}_1,\cdots,\tilde{\boldsymbol{c}}_M$, which then attend back with input $f_1,\cdots,f_N$ in Transformer 2 to generate the output features $\tilde{f}_1, \cdots, \tilde{f}_N$. (b) Zoom-in of Transformer 1. $\boldsymbol{K}^s$  and $\boldsymbol{K}^t$ are the key projections used on the source and target domains, respectively.} \label{arch}
\end{center}
\end{figure}

To bridge the domain gap, a polyformer layer is inserted between $M_1$ and $M_2$, as shown in Fig. \ref{pipeline}(b). Now $M_2$ processes transformed features $\boldsymbol{\tilde{f}}$. After training the polyformer layer on the source domain and  fine-tuning it on the target domain, $\boldsymbol{\tilde{f}}$ should ``look more familiar'' to $M_2$, leading to improved prediction accuracy.
Instead of seeking adaptation between the myriads of features in the source and target domains, we propose to first find a condensed representation of the source-domain features, namely a set of \emph{prototype embeddings} (prototypes). Adaptation becomes much easier on a much smaller set of prototypes, making the model suited for few-shot scenarios. 

\subsection{Polyformer Architecture}
Different designs can be chosen to implement the polyformer layer, and here we adopt the Squeeze-and-Expansion Transformer proposed in \cite{segtran}. Fig.\ref{arch} presents (a) the schematic of the two-step attention by the two sub-transformers, and (b) the zoomed-in architecture, especially the structure of Transformer 1.

A polyformer layer consists of two sub-transformer layers, denoted as Transformer 1 and 2. The prototypes are a set of $M$ persistent embeddings $\boldsymbol{c}_1,\cdots,\boldsymbol{c}_M$, independent of the input. The input features $\boldsymbol{f}$ consist of a set of feature vectors $\{\boldsymbol{f}_1,\cdots,\boldsymbol{f}_N\}$, which are obtained by spatially flattening the input feature maps. Transformer 1 performs cross-attention between the prototypes and the input features, yielding intermediate features $\tilde{\boldsymbol{C}} = \tilde{\boldsymbol{c}}_1,\cdots,\tilde{\boldsymbol{c}}_M$. They attend back with $\boldsymbol{f}$ in Transformer 2, outputting the adapted features $\boldsymbol{\tilde{f}} = \{ \tilde{\boldsymbol{f}}_1,\cdots,\tilde{\boldsymbol{f}}_N \}$: 
\begin{align}
\tilde{\boldsymbol{C}} &= \text{Transformer1}(\boldsymbol{f}, \boldsymbol{C}), \\
\boldsymbol{\tilde{f}} &= \text{Transformer2}(\tilde{\boldsymbol{C}}, \boldsymbol{f}) + \boldsymbol{f}, \label{transformer2}
\end{align}
where in Eq.\eqref{transformer2}, a residual connection builds upon the original representations. In a Squeeze-and-Expansion transformer, sub-transformers 1 and 2 are two Expanded Attention Blocks \cite{segtran}. In each sub-transformer, the attention matrix is of $N\times M$. Typically, the number of feature points $N > 10^4$, and we chose $M=256$. Thus, the huge feature space is compressed using a set of 256 prototypes.

For the purpose of domain adaptation, our focus is on Transformer 1:
\begin{align}
\operatorname{Att\_weight}(\boldsymbol{f},\boldsymbol{C}) & = \sigma(\boldsymbol{K}(\boldsymbol{f}), \boldsymbol{Q}(\boldsymbol{C})) \in \mathbb{R}^{N \times M}, \label{att-weight} \\
 \operatorname{Attention}(\boldsymbol{f},\boldsymbol{C}) &= \operatorname{Att\_weight}(\boldsymbol{f},\boldsymbol{C}) \cdot \boldsymbol{V}(\boldsymbol{f}), \label{attention-eqn} \\ 
\tilde{\boldsymbol{C}} &= \operatorname{FFN}(\operatorname{Attention}(\boldsymbol{f},\boldsymbol{C})),
\end{align}
where $\boldsymbol{K}, \boldsymbol{Q}, \boldsymbol{V}$ are key, query, and value projections, respectively. $\sigma$ is softmax after dot product. $\operatorname{Att\_weight}(\boldsymbol{f},\boldsymbol{C})$ is a pairwise attention matrix, whose $i,j$-th element defines how much $\boldsymbol{f}_i$ contributes to the fused features of prototype $j$.  $\operatorname{FFN}$ is a feed-forward network that transforms the fused features into $\tilde{\boldsymbol{c}}_j$. 

In Eq.(\ref{att-weight}), $\boldsymbol{K}$ controls which subspace input features $f_i$ are projected to, and influences the attention between $\boldsymbol{f}$ and $\boldsymbol{C}$. It leads to the following hypothesis:
\begin{hypothesis}
 In the target domain, by properly updating $\boldsymbol{K}: \boldsymbol{K}^s \to \boldsymbol{K}^t$, the polyformer layer will project input features $\boldsymbol{f}$ (which follow a different distribution) to a subspace similar to that in the source domain. Consequently, the output features from the polyformer are semantically compatible with the target-domain output feature space. As a result, the model will perform better on the target domain without updating the original model weights.
\end{hypothesis}

According to Hypothesis 1, to adapt a trained model on a new domain, we can share $\boldsymbol{Q}, \boldsymbol{V}$ and FFN across domains, and update $\boldsymbol{K}$ only. This scheme inherites most of the representation powers from the source domain. 

When the polyformer layer is an ordinary transformer without prototypes, the attention matrix is a huge ${N \times N}$ matrix ($N>10^4$). Then cross-domain semantic alignment becomes much more harder, and hypothesis 1 may not satisfy.

\subsection{Training and Adaptation of the Polyformer Layer}
\subsubsection{Training on the Source Domain.}
After a polyformer layer is inserted in a model $M$ trained on the source domain, we need to train the polyformer layer to make the new pipeline $M'$ keep similar performance on the source domain as the original pipeline. This is achieved by training again on the source-domain data. 
Specifically, all the weights (including BatchNorms) of $M$ are frozen, and only the polyformer weights are to be updated. The same training protocol is performed on the source-domain training data $\{(x^s_1, y^s_1), \cdots, (x^s_n, y^s_n)\}$. After training, the prototypes are compressed representations of the source domain features.

\subsubsection{Adapting to the Target Domain.}
On the target domain, all the weights (excluding BatchNorms) of $M$ are frozen, and only the $\boldsymbol{K}$ projection weights and BatchNorm parameters are to be updated. The training is performed on the few-shot target-domain training data $(X^t, Y^t)=\{(x^t_1, y^t_1), \cdots, (x^t_m, y^t_m)\}$. 

Note that traditional domain adversarial losses \cite{revgrad,adda} could be incorporated to improve the adaptation, as shown in our ablation studies (Section \ref{ablation}):
\begin{equation}
\mathcal{L}_{adapt}(X^s, X^t, Y^t) = \mathcal{L}_{sup}(X^t, Y^t) + \mathcal{L}_{adv}(X^s, X^t).
\end{equation}
There are two common choices for the domain adversarial loss: the discriminator could try to discriminate either 1) the features of a source vs. a target domain image, or 2) the predicted masks on a a source vs. a target domain image.

\section{Experiments}
Different methods were evaluated on two medical image segmentation tasks:
\paragraph{Optic Disc/Cup Segmentation.}
This task does segmentation of the optic disc and cup in fundus images, which are 2D images of the rear of the eyes. The source domain was the 1200 training images provided in the REFUGE  challenge~\cite{refuge}. The target domain, the RIM-One dataset~\cite{rim-one}, contains 159 images.

\paragraph{Polyp Segmentation.}
This task does polyp (fleshy growths inside the colon lining) segmentation in colonoscopy images. The source domain was a combination of two datasets: CVC-612 (612 images)~\cite{7840040} and Kvasir (1000 images)~\cite{Pogorelov}. The target domain was the CVC-300 dataset (60 images)~\cite{pranet}.

\vspace{0.5em}\noindent\textbf{Number of shots.}\hspace{0.5em} For each task, \textbf{five} annotated images were randomly selected from the target domain to do few-shot supervised training. Each method was evaluated on the remaining target-domain images. Results with 10, 15 and 20 shots can be found in the supplementary file.

\subsection{Ablation Studies} \label{ablation}
\begin{table}[t]
\begin{centering}
\begin{tabular}{|P{5.5cm}|P{1.3cm}|P{1.3cm}|P{1.6cm}|P{1.2cm}|}\hline
& \multicolumn{2}{c|}{RIM-One} & \multirow{2}{*}{CVC-300} & \multirow{2}{*}{Avg.} \tabularnewline
\cline{2-3} 
 & Disc & Cup & &\tabularnewline
\hline 
\textbf{Trained on Source Domain} & \multicolumn{4}{c|}{} \tabularnewline \hline 
U-Net & 0.819 & 0.708 & 0.728 & 0.752 \tabularnewline \hline 
Polyformer & 0.815 & 0.717 & 0.724 & 0.752 \tabularnewline \hline 
\hhline{|=|=|=|=|=|}
\textbf{Adapted to Target Domain} & \multicolumn{4}{c|}{} \tabularnewline \hline 
$\mathcal{L}_{adv}+K$         & 0.828 & 0.731 & 0.779 & 0.779 \tabularnewline \hline 
$\mathcal{L}_{sup}+K$, w/o BN & 0.823 & 0.741 & 0.760 & 0.775 \tabularnewline \hline 
$\mathcal{L}_{sup}+K$         & 0.900 & 0.753 & 0.830 & 0.828 \tabularnewline \hline 
$\mathcal{L}_{sup}+\mathcal{L}_{adv}$ + All weights & 0.892 & 0.741 & 0.826 & 0.820 \tabularnewline \hline 
$\mathcal{L}_{sup}+\mathcal{L}_{adv}\text{(mask)}+K$ & 0.909 & \textbf{0.763} & \textbf{0.836} & \textbf{0.836} \tabularnewline \hline 
$\mathcal{L}_{sup}+\mathcal{L}_{adv}$ +$K$ (standard setting) & \textbf{0.913} & 0.758 & 0.834 & 0.835 \tabularnewline \hline 
\end{tabular}\vspace{2pt}
\caption{The dice scores on Fundus and Polyp target domains RIM-One and CVC-300, by five ablated Polyformer models and the standard  ``$\mathcal{L}_{sup}+\mathcal{L}_{adv}$ +$K$''.
The U-Net and Polyformer trained on the source-domain were includes
as references.}
\label{ablation-scores}
\par\end{centering}
\end{table}

A standard Polyformer and five ablations were evaluated on the two tasks:  
\begin{itemize}
\item $\mathcal{L}_{sup}+\mathcal{L}_{adv}$ +$K$ \textbf{(standard setting)}, i.e.,
fine-tuning only the $\boldsymbol{K}$ projection, with both the few-shot supervision and the domain adversarial learning on features. It is the standard setting from which other ablations are derived;
\item $\mathcal{L}_{adv}+K$, i.e., fine-tuning the $\boldsymbol{K}$ projection
using the unsupervised domain adversarial loss on features, without using the few-shot supervision; 
\item $\mathcal{L}_{sup}+K$, w/o BN, i.e., freezing the BatchNorm affine
parameters, but still updating the mean/std statistics on the target
domain; 
\item $\mathcal{L}_{sup}+K$, i.e., fine-tuning the $\boldsymbol{K}$ projection using the few-shot supervision only, without the domain adversarial loss; 
\item $\mathcal{L}_{sup}+\mathcal{L}_{adv}$ + All weights, i.e., fine-tuning
the whole polyformer layer, instead of only the $\boldsymbol{K}$
projection; 
\item $\mathcal{L}_{sup}+\mathcal{L}_{adv}\text{{(mask)}}+K$, i.e., doing
domain adversarial learning on the predicted masks, 
instead of on the extracted features. 
\end{itemize}

Table \ref{ablation-scores} presents the results of the standard setting ``$\mathcal{L}_{sup}+\mathcal{L}_{adv}$ +$K$'', as well as five ablated models.
 Without the few-shot supervision, the domain adversarial loss only marginally improved the target-domain performance ($0.752 \to 0.779$). Freezing the BatchNorm affine parameters greatly restricts adaptation ($0.752 \to 0.775$).
Fine-tuning the whole polyformer layer led to worse performance than fine-tuning the $\boldsymbol{K}$ projection only (0.820 vs. 0.835), probably due to catastrophic forgetting \cite{no-forget} of the source-domain semantics encoded in the prototypes. Incorporating the domain adversarial complemented and helped the few-shot supervision obtain better performance ($0.828 \to 0.835$). The domain adversarial loss on features led to almost the same results as on masks.

\subsection{Compared Methods}
Two settings of Polyformer, as well as ten popular baselines, were evaluated:
\begin{itemize}
    \item \textbf{U-Net (source)}, trained on the source domain without adaptation;
    \item \textbf{$\mathcal{L}_{sup}$}, fine-tuning U-Net (source) on the five target-domain images;
    \item \textbf{$\mathcal{L}_{sup}$(source + target)}, trained on a mixture of all source-domain images and the five target-domain images;
    \item \textbf{CycleGAN + $\mathcal{L}_{sup}$(source)} \cite{cyclegan,dual-teacher}\footnote{CycleGAN is the core component for DA in \cite{dual-teacher}, but \cite{dual-teacher} is more than CycleGAN.}. The CycleGAN was trained for 200 epochs to convert between the source and the target domains. The converted source-domain images were used to train a U-Net from scratch;
    \item \textbf{RevGrad ($\mathcal{L}_{sup} + \mathcal{L}_{adv}$)} \cite{revgrad}, which fine-tunes U-Net (source), by optimizing the domain adversarial loss on the features with a gradient reversal layer;
    \item \textbf{ADDA ($\mathcal{L}_{sup} + \mathcal{L}_{adv}$)} \cite{adda}, which uses inverted domain labels to replace the gradient reversal layer in RevGrad for more stable gradients; 
    \item \textbf{DA-ADV (tune whole model)} \cite{da-adv} also named as $p$OSAL in \cite{posal}, which fine-tunes the whole U-Net (source) by discriminating whether the masks are generated on the source or the target domain images using RevGrad;
    \item \textbf{DA-ADV (tune last two layers)}, DA-ADV training that only fine-tunes the last two layers and all BatchNorm parameters of U-Net (source);
    \item \textbf{CellSegSSDA ($\mathcal{L}_{sup}$+$\mathcal{L}_{adv}$\textnormal{(mask)}+$\mathcal{L}_{recon}$)} \cite{adv-adapt-cell}, which combines RevGrad on predicted masks, an image reconstruction loss and few-shot supervision; %The same hyperparameters $\lambda_{adv}=0.001$ and $\lambda_{recons}=0.01$ were adopted.
    \item \textbf{Polyformer ($\mathcal{L}_{sup}$ +$K$)}, by fine-tuning the $\boldsymbol{K}$ projection in the polyformer layer, with the few-shot supervision only;
    \item \textbf{Polyformer ($\mathcal{L}_{sup} + \mathcal{L}_{adv}$ +$K$)}, i.e., the standard setting of Polyformer training, which enhances Polyformer ($\mathcal{L}_{sup}$ +$K$) with RevGrad on features;
    \item \textbf{$\mathcal{L}_{sup}$ (50\% target)}, by fine-tuning U-Net (source) on 1/2 of the target-domain images, and tested on the remaining 1/2 images. This serves as an empirical upper-bound of all methods\footnote{However, the performance of $\mathcal{L}_{sup}$ (50\% target) on CVC-300 was lower than Polyformer and other baseline methods with more shots, partly because CVC-300 is small (60 images) and sensitive to randomness. See the supplementary file for discussions and more experiments.}.
\end{itemize}
The domain adversarial methods RevGrad, ADDA and DA-ADV were combined with the few-shot supervision to do semi-supervised learning.
All the methods were trained with a batch size of 4, and optimized with the AdamW optimizer at an initial learning rate of 0.001. The supervised training loss was the average of the pixel-wise cross-entropy loss and the dice loss. 

\begin{table}[t]
\begin{centering}
\begin{tabular}{|P{6.6cm}|P{1.2cm}|P{1.2cm}|P{1.4cm}|P{1.1cm}|}\hline
& \multicolumn{2}{c|}{RIM-One} & \multirow{2}{*}{CVC-300} & \multirow{2}{*}{Avg.} \tabularnewline
\cline{2-3} 
 & Disc & Cup & &\tabularnewline
\hline 
U-Net (source) \cite{unet} & 0.819 & 0.708 & 0.728 & 0.752 \tabularnewline
\hline 
$\mathcal{L}_{sup}$ \cite{unet} & 0.871 & 0.665 & 0.791 & 0.776 \tabularnewline
\hline 
$\mathcal{L}_{sup}$(source + target) \cite{unet} & 0.831 & 0.715 & 0.808 & 0.785 \tabularnewline
\hline 
CycleGAN + $\mathcal{L}_{sup}$(source) \cite{cyclegan,dual-teacher} & 0.747 & 0.690 & 0.709 & 0.715 \tabularnewline
\hline 
RevGrad ($\mathcal{L}_{sup} + \mathcal{L}_{adv}$) \cite{revgrad} & 0.860 & 0.732 & 0.813 & 0.802 \tabularnewline
\hline 
ADDA ($\mathcal{L}_{sup} + \mathcal{L}_{adv}$) \cite{adda} & 0.874 & 0.726 & \textbf{0.836} & 0.812 \tabularnewline
\hline 
DA-ADV (tune whole model) \cite{da-adv,posal} & 0.885 & 0.725 & 0.830 & 0.813 \tabularnewline
\hline 
DA-ADV (tune last two layers) \cite{da-adv,posal} & 0.872 & 0.730 & 0.786 & 0.796 \tabularnewline
\hline 
CellSegSSDA ($\mathcal{L}_{sup}$+$\mathcal{L}_{adv}$\textnormal{(mask)}+$\mathcal{L}_{recon}$) \cite{adv-adapt-cell} & 0.869 & 0.756 & 0.805 & 0.810 \tabularnewline
\hline 
Polyformer ($\mathcal{L}_{sup}$ +$K$) & 0.900 & 0.753 & 0.830 & 0.828 \tabularnewline \hline 
Polyformer ($\mathcal{L}_{sup}+\mathcal{L}_{adv}$ +$K$) & \textbf{0.913} & \textbf{0.758} & 0.834 & \textbf{0.835} \tabularnewline \hhline{|=|=|=|=|=|}
%$\mathcal{L}_{sup}$ (50\% target) \cite{unet} & 0.959 & 0.834 & 0.882 & 0.892 
$\mathcal{L}_{sup}$ (50\% target) \cite{unet} & 0.959 & 0.834 & 0.834 & 0.876
\tabularnewline \hline 
\end{tabular}\vspace{2pt}
\caption{Dice scores on Fundus and Polyp target domains RIM-One and CVC-300.}
\label{scores}
\par\end{centering}
\end{table}

\subsection{Results}
Table \ref{scores} presents the segmentation performance of different methods on the two target domains, measured in dice scores. The domain adversarial loss effectively reduced the performance gap between the source and target domains. ``CycleGAN + $\mathcal{L}_{sup}$(source)'' performed even worse than U-Net (source), as CycleGAN does not guarantee semantic alignment when doing conversion \cite{cyclegan}. 
Without the domain adversarial loss, Polyformer ($\mathcal{L}_{sup}$ +$K$) has already achieved higher average dice scores than all the baseline methods. By incorporating the domain adversarial loss RevGrad on predicted masks, Polyformer ($\mathcal{L}_{sup} + \mathcal{L}_{adv}$ +$K$) achieved higher performance than Polyformer ($\mathcal{L}_{sup}$ +$K$), showing that Polyformer is complementary with the traditional domain adversarial loss. 
%Although overall, the empirical upper-bound method $\mathcal{L}_{sup}$ (50\% target) performed much better thanks to many more training images, it performed slightly worse than Polyformer ($\mathcal{L}_{sup} + \mathcal{L}_{adv}$) on the disc segmentation of fundus images. This might be caused by the competition between the source and target domain for the model capacity. Similar results were also observed in \cite{adv-adapt-cell}.

\begin{figure}[t]
\centering
  \includegraphics[scale=0.34]{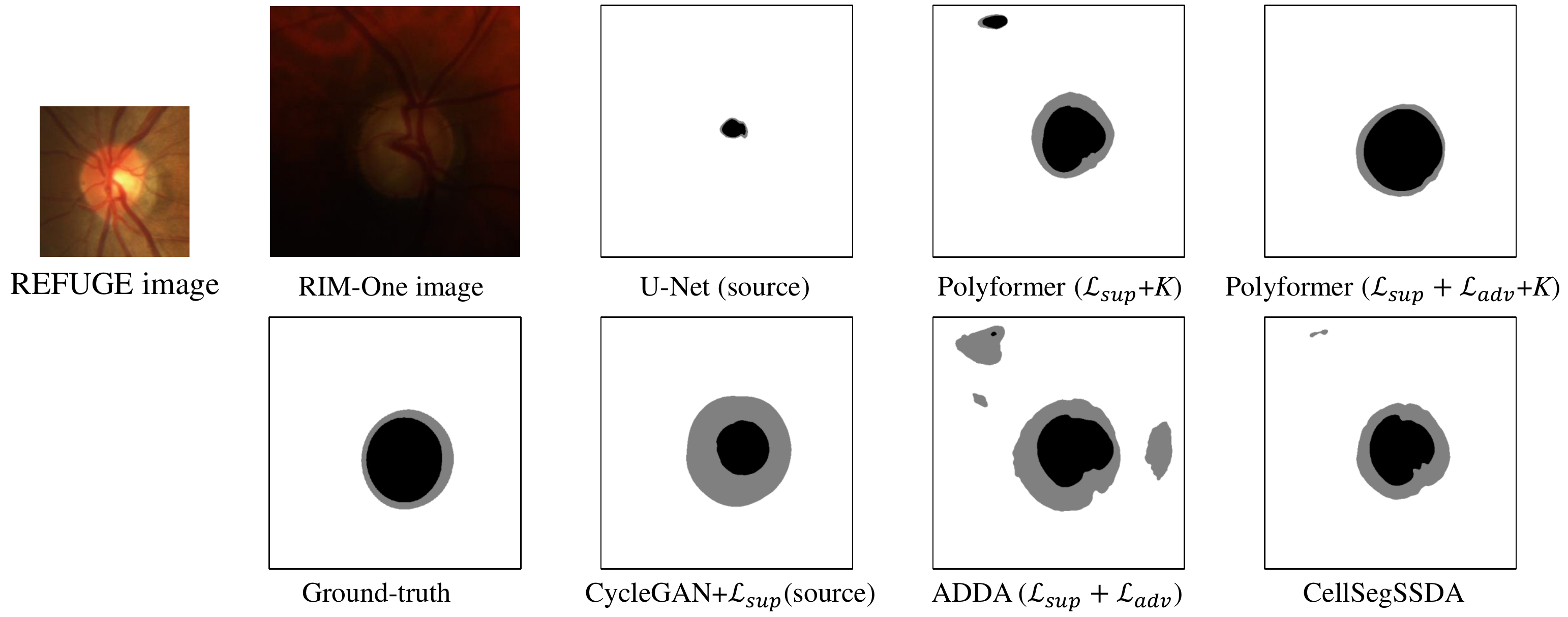}
  \captionof{figure}{The segmentation masks predicted by different methods on a RIM-One image.}
  \label{fig:rim-one}
\end{figure}

To gain an intuitive understanding of how different methods performed, Fig. \ref{fig:rim-one} presents a fundus image from RIM-One, the ground-truth segmentation mask and the predicted masks by selected methods. In addition, a REFUGE image is presented in the left to visualize the source/target domain gap. Without adaptation, U-Net (source) was unable to find most of the disc/cup areas, as the image is much darker than typical source-domain REFUGE images. The mask predicted by ``CycleGAN + $\mathcal{L}_{sup}$(source)'' largely deviates from the ground-truth. The mask from Polyformer was significantly improved by the domain adversarial loss, in that the artifacts were eliminated and the mask became much closer to the ground-truth. For comparison purposes, another example is presented in the supplementary file where all the major methods failed.

\section{Conclusions}
In this work, we proposed a plug-and-play module called the polymorphic transformer (Polyformer) for domain adaptation. It can be plugged into a pretrained model. On a new domain, only fine-tuning a projection layer within Polyformer is sufficient to achieve good performance. 
We demonstrated the effectiveness of Polyformer on two segmentation tasks, where it performed consistently better than strong baselines in the challenging few-shot learning setting.

\section*{Acknowledgements}
We are grateful for the help and support of Wei Jing. This research is supported by A*STAR under its Career Development Award (Grant No. C210112016), and its Human-Robot Collaborative Al for Advanced Manufacturing and Engineering (AME) programme (Grant No. A18A2b0046).

\bibliographystyle{splncs04}
\bibliography{polyformer}

\ifx\hasappendix\true
\clearpage
\appendix
\section{Architecture of U-Net and Discriminator}
All baseline methods and the polyformer are based on a popular U-Net implementation  \url{https://github.com/milesial/Pytorch-UNet}, which has four downsampling and four upsampling layers. 

The discriminator is 5-layer CNN, with a gradient-reversal layer inserted before all the layers. 

\section{Hyperparameters of Polyformer}
There are two important hyperparameters of polyformer: the number of prototypes $M$, and the number of modes $N_m$ of the extended attention block \cite{segtran}. Our experiments show that, when reducing $M$ from the default 256 to 128 or 64, the performance only dropped very slightly. $N_m$ has bigger impact to the model performance. Typically $N_m=4$ or 2. On RIM-One, $N_m=4$ performs slightly better than $N_m=2$. On  CVC-300, $N_m=2$ performs slightly better instead.

\section{Results of More Shots}
\begin{table}[hbtp]
\begin{centering}
\begin{tabular}{|P{6.6cm}|P{1.2cm}|P{1.2cm}|P{1.4cm}|P{1.1cm}|}\hline
& \multicolumn{2}{c|}{RIM-One} & \multirow{2}{*}{CVC-300} & \multirow{2}{*}{Avg.} \tabularnewline
\cline{2-3} 
 & Disc & Cup & &\tabularnewline
\hline 
\textbf{5-Shot} & \multicolumn{4}{c|}{} \tabularnewline \hline 
$\mathcal{L}_{sup}$ \cite{unet} & 0.871 & 0.665 & 0.791 & 0.776 \tabularnewline
\hline 
RevGrad ($\mathcal{L}_{sup} + \mathcal{L}_{adv}$) \cite{revgrad} & 0.860 & 0.732 & 0.813 & 0.802 \tabularnewline
\hline 
CellSegSSDA ($\mathcal{L}_{sup}$+$\mathcal{L}_{adv}$\textnormal{(mask)}+$\mathcal{L}_{recon}$) \cite{adv-adapt-cell} & 0.869 & 0.756 & 0.805 & 0.810 \tabularnewline
\hline 
Polyformer ($\mathcal{L}_{sup} + \mathcal{L}_{adv}$) & \textbf{0.913} & \textbf{0.758} & \textbf{0.834} & \textbf{0.835} \tabularnewline \hline 

\textbf{10-Shot} & \multicolumn{4}{c|}{} \tabularnewline \hline 
$\mathcal{L}_{sup}$ \cite{unet} & 0.924 & 0.744 & 0.819 & 0.829 \tabularnewline
\hline 
RevGrad ($\mathcal{L}_{sup} + \mathcal{L}_{adv}$) \cite{revgrad} & 0.877 & 0.744 & 0.835 & 0.819 \tabularnewline
\hline 
CellSegSSDA $\mathcal{L}_{sup}$+$\mathcal{L}_{adv}$\textnormal{(mask)}+$\mathcal{L}_{recon}$) \cite{adv-adapt-cell} & 0.918 & 0.722 & \textbf{0.860} & 0.833 \tabularnewline
\hline 
Polyformer ($\mathcal{L}_{sup} + \mathcal{L}_{adv}$ +$K$) & 0.915 & \textbf{0.767} & 0.844 & \textbf{0.842} \tabularnewline \hline 

\textbf{15-Shot} & \multicolumn{4}{c|}{} \tabularnewline \hline 
$\mathcal{L}_{sup}$ \cite{unet} & 0.933 & 0.800 & 0.799 & 0.844 \tabularnewline
\hline 
RevGrad ($\mathcal{L}_{sup} + \mathcal{L}_{adv}$) \cite{revgrad} & 0.906 & 0.765 & 0.854 & 0.842 \tabularnewline
\hline 
CellSegSSDA ($\mathcal{L}_{sup}$+$\mathcal{L}_{adv}$\textnormal{(mask)}+$\mathcal{L}_{recon}$) \cite{adv-adapt-cell} & 0.911 & 0.775 & \textbf{0.874} & 0.853 \tabularnewline
\hline 
Polyformer ($\mathcal{L}_{sup} + \mathcal{L}_{adv}$ +$K$) & 0.925 & \textbf{0.801} & 0.853 & \textbf{0.860} \tabularnewline \hline 

\textbf{20-Shot} & \multicolumn{4}{c|}{} \tabularnewline \hline 
$\mathcal{L}_{sup}$ \cite{unet} & \textbf{0.937} & \textbf{0.824} & 0.820 & 0.860 \tabularnewline
\hline 
RevGrad ($\mathcal{L}_{sup} + \mathcal{L}_{adv}$) \cite{revgrad} & 0.911 & 0.806 & 0.858 & 0.858 \tabularnewline
\hline 
CellSegSSDA ($\mathcal{L}_{sup}$+$\mathcal{L}_{adv}$\textnormal{(mask)}+$\mathcal{L}_{recon}$) \cite{adv-adapt-cell} & 0.931 & 0.801 & \textbf{0.872} & \textbf{0.868} \tabularnewline
\hline 
Polyformer ($\mathcal{L}_{sup} + \mathcal{L}_{adv}$ +$K$) & 0.927 & 0.813 & 0.860 & 0.867 \tabularnewline \hhline{|=|=|=|=|=|}
$\mathcal{L}_{sup}$ (50\% target) \cite{unet} & 0.959 & 0.834 & 0.834 & 0.876 \tabularnewline \hline 
\end{tabular}\vspace{2pt}
\caption{Dice scores on Fundus and Polyp target domains RIM-One and CVC-300.}
\label{more-shots}
\par\end{centering}
\end{table}

Table \ref{more-shots} presents the results of 10, 15 and 20 shots on polyformer and a few representative baselines. In these settings, polyformer still outperformed the baseline methods.

It is somewhat surprising that, with more training examples on CVC-300 (a polyp segmentation dataset), the three methods with the domain adversarial loss still outperformed $\mathcal{L}_{sup}$ (50\% target) by 3-5\%. It shows that the domain adversarial loss is important for transferring the learned representations from the source domain to new domains.

One thing to note is that, as the dataset CVC-300 is small (60 images), the evaluation results on it are sensitive to various randomness, including the train/test data split, model initialization, the types of adopted model regularizations, and other hyperparameters. For the experiments we have controlled the train/test data split and common hyperparameters to ensure different methods were evaluated on the same ground. However, this may still be not totally fair for some baselines. For example, it is possible that under a carefully-chosen  learning rate, $\mathcal{L}_{sup}$ may perform significantly better. 

In the future, we would like to evaluate different methods on bigger datasets, so that more reliable conclusions would be drawn from the experimental results. 

\section{A Failed Example}
Fig. \ref{fig:rim-one-bad} presents a difficult RIM-One image on which all the compared methods failed. Different from Fig. 3 in the main paper where the domain adversarial loss $\mathcal{L}_{adv}$ helps remove artifacts, here $\mathcal{L}_{adv}$ introduces two ``phantom optic discs/cups'' to the mask produced by the polyformer. A possible explanation is that the discriminator is implemented as a Convolutional Neural Network (CNN), and due to the local nature of CNNs, the regularization from it is local, i.e., effective within small neighborhoods. Thus, after seeing a highly-likely optic disc/cup area (i.e., the groundtruth area), it could not suppress other ``phantom optic discs/cups''. Instead, it just refines each specious optic disc/cup area, so that when examined locally, each area looks like a more reasonable optic disc/cup; but in the end, it makes the whole mask incongruous and illogical.

For future work, we may improve the discriminator by incorporating a transformer, so that it will provide regularization at a more global level.

\begin{figure}[hbtp]
\centering
  \includegraphics[scale=0.34]{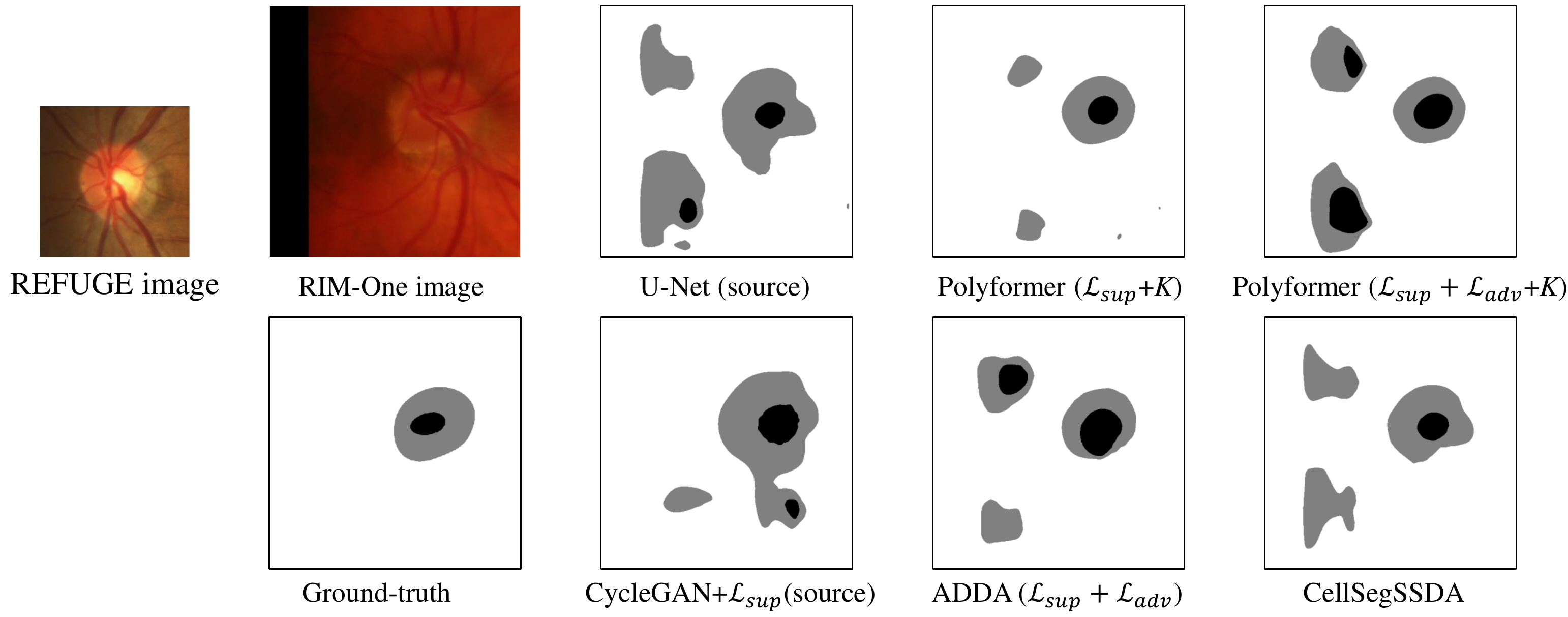}
  \captionof{figure}{A difficult RIM-One image on which all methods failed.}
  \label{fig:rim-one-bad}
\end{figure}
\fi

\end{document}